%% file: CF4BPM2021.tex
\documentclass[runningheads]{llncs}
\usepackage{graphicx}
\usepackage{xcolor}
\usepackage{bbm}
\usepackage{multirow}
\usepackage{booktabs}
\usepackage{colortbl}
\usepackage{caption}
\usepackage{subcaption}
\usepackage{amsmath}
\usepackage{hyperref}
\usepackage{url}
\usepackage{wrapfig}

\usepackage{xcolor,soul}
\sethlcolor{lightgray}

\newcommand{\runin}[1]{\vspace{.2em}\noindent\textbf{#1}}

\newcommand{\loreley}{{\sc loreley}}
\newcommand{\lime}{{\sc lime}}
\newcommand{\lore}{{\sc lore}}

\begin{document}
\title{Counterfactual Explanations for Predictive Business Process Monitoring}
%
%
\author{
    Tsung-Hao Huang \and
    Andreas Metzger \and
    Klaus Pohl
    }
\authorrunning{T. Huang et al.}
%
\institute{paluno - The Ruhr Institute for Software Technology \\
Gerlingstraße 16, 45127 Essen, Germany
\\
\email{\{tsunghao.huang, andreas.metzger, klaus.pohl\}@paluno.uni-due.de}}

\maketitle              
\begin{abstract}
Predictive business process monitoring increasingly leverages sophisticated prediction models.
Although sophisticated models achieve consistently higher prediction accuracy than simple models, one major drawback is their lack of interpretability, which limits their adoption in practice. 
We thus see growing interest in explainable predictive business process monitoring, which aims to increase the interpretability of prediction models. 
Existing solutions focus on giving factual explanations.
While factual explanations can be helpful, humans typically do not ask why a particular prediction was made, but rather why it was made instead of another prediction, i.e., humans are interested in counterfactual explanations.
While research in explainable AI produced several promising techniques to generate counterfactual explanations, directly applying them to predictive process monitoring may deliver unrealistic explanations, because they ignore the underlying process constraints.
We propose LORELEY, a counterfactual explanation technique for predictive process monitoring, which extends LORE, a recent explainable AI technique. 
We impose control flow constraints to the explanation generation process to ensure realistic counterfactual explanations. 
Moreover, we extend LORE to enable explaining multi-class classification models.
Experimental results using a real, public dataset indicate that LORELEY can approximate the prediction models with an average fidelity of  97.69\% and generate realistic counterfactual explanations.


\end{abstract}
\input{intro}

\input{sota}

\input{prob}

\input{approach}

\input{evaluation}

\input{limitations}

\section{Conclusion and Outlook}
This paper introduced \loreley{}, a counterfactual explanation technique for process predictions.
\loreley{} extends \lore{} \cite{DBLP:journals/expert/GuidottiMGPRT19} to ensure the explanations conform to the underlying process constraints.
Experimental results 
indicate that 
\loreley{} is able to approximate the prediction models with high fidelity and generate realistic counterfactual explanations.

As future work, we plan to address the current limitations of the technique.
In addition, we will evaluate \loreley\ with further benchmark datasets and different black-box prediction models, as well as analyze the stability of the technique by comparing the difference of explanations generated by different runs for the same input. 

\vspace{1em}\footnotesize{\textbf{Acknowledgments.} 
We thank Felix Feit, Tristan Kley, and Xhulja Shahini for comments on earlier drafts.
Research leading to these results received funding from the European Union’s Horizon 2020 R\&I programme under grant  no. 871493 (DataPorts).}

\bibliographystyle{splncs04}
\bibliography{myrefs}






\end{document}

%% file: intro.tex
\section{Introduction}
\label{sec:intro}

Predictive business process monitoring forecasts how an ongoing business process instance (aka. case) will unfold by utilizing prediction models trained from historical process data (typically in the form of event logs)~\cite{Francescomarino18,Marquez-Chamorro_2017}.
One may, e.g., predict the remaining time~\cite{DBLP:journals/tist/VerenichDRMT19}, outcome of an ongoing case~\cite{TeinemaaDRM19} or the next activity executed~\cite{DBLP:journals/sosym/TaxTZ20}.

Increasingly, sophisticated prediction models are employed for predictive process monitoring.
These sophisticated models include tree ensembles \cite{TeinemaaDRM19,verenich2016complex} and deep artificial neural networks \cite{EvermannRF17,ParkS20,TaxVRD17}.
Compared to simple prediction models, such as decision trees or linear regression, these sophisticated prediction models achieve consistently better accuracy in various types of predictive process monitoring problems~\cite{DBLP:journals/sosym/TaxTZ20,TaxVRD17,DBLP:journals/tist/VerenichDRMT19}.

However, one drawback of these sophisticated prediction models is their lack of interpretability.
Using black-box prediction models without being able to interpret their decisions has potential risks \cite{guidotti2018survey,Miller19,ribeiro2016should}.

To facilitate the interpretability of black-box prediction models, research under the name of explainable AI is gaining interest~\cite{guidotti2018survey,Miller19}.
Consequently, explainable AI techniques have started to be applied also in the context of predictive process monitoring, helping to deliver accurate predictions which can be interpreted.
Existing explainable predictive process monitoring techniques give explanations for why a particular prediction was made.
While factual explanations can be helpful, people typically do not ask why a particular prediction was made, but rather why it was made instead of another prediction
\cite{Miller19,molnar2019}. 
Counterfactual explanations help people to reason on cause-effect or reason-action relationships between events~\cite{Byrne19,Miller19}.

We introduce \loreley{}, a technique to generate counterfactual explanations for process predictions.
\loreley\ is based on the model-agnostic technique \lore\ introduced by Guidotti et al.~\cite{DBLP:journals/expert/GuidottiMGPRT19}.
\lore\ generates explanations using genetic algorithms.
We demonstrate the feasibility of \loreley{} using a real and public dataset, the BPIC2017 event log.
Results indicate that we can reach a fidelity of the explanations of 97.69\%, i.e., \loreley{} can very accurately and faithfully approximate the local decision boundary of the black-box prediction model.
In addition, \loreley{} generates high-quality explanations, matching the domain knowledge about the process.

Sect.~\ref{sec:sota} gives a discussion of related work.
Sect.~\ref{sec:rationale} provides a detailed problem statement.
Sect.~\ref{sec:approach} introduces the \loreley\ technique.
Sect.~\ref{sec:Experiments} describes the setup and Sect.~\ref{sec:results} the results of our evaluation.
Sect.~\ref{sec:limit} discusses current limitations.

%% file: sota.tex
\section{Related Work}
\label{sec:sota}

We structure related work on explainable predictive process monitoring along with two types of techniques: \emph{model-specific} and \emph{model-agnostic}~\cite{guidotti2018survey,RibeiroSG16a}.

\runin{Model-specific Explanation Techniques.}
Model-specific techniques are tied to a specific type of prediction model.
One direction of work uses attribute importance derived from neural networks as an explanation.
Weinzierl et al. use layer-wise relevance propagation to derive attribute importance from LSTMs~\cite{WeinzierlZBRM020}. 
Sindhgatta et al. use the attention mechanism of LSTMs to extract attribute importance~\cite{SindhgattaMOB20}. 
Harl et al. apply gated graph neural networks and extract attribute importance from the softmax layer for each event in a process instance~\cite{harl2020explainable}. 

Another direction of work suggests designing transparent models, which are interpretable by users and at the same time provide accurate predictions.
Breuker et al. propose combining process mining and grammatical inference~\cite{BreukerMDB16}. 
Their solution consists of a predictor and an analyzer. While the predictor provides the predictions using a probabilistic finite automaton, the analyzer provides visualizations of the predictor.  
Verenich et al. propose an interpretable way to predict the remaining cycle time of a process instance by building on a mined process model~\cite{verenich2019predicting}.  
For each activity in the process model, a regression model is trained to predict its execution time, and for each decision point, a classification model is trained 
to predict branching probabilities. 
B{\"o}hmer and Rinderle-Ma propose sequential prediction rules to predict the temporal behavior and the next activity of an ongoing case~\cite{bohmer2020logo}.  

The above work demonstrates the feasibility and relevance of explainable predictive process monitoring but focuses on factual explanations only.
Our technique complements this work by generating counterfactual explanations. 
In addition, our technique is model-agnostic and thus -- compared to the above work -- can generate explanations for any type of prediction model.
One benefit is that the form of explanations remains the same even if the underlying prediction model changes, and thus model-agnostic techniques facilitate evaluating alternatives among different prediction models before their deployment \cite{RibeiroSG16a}.

\runin{Model-agnostic Explanation Techniques.}
Model-agnostic techniques can generate explanations for any type of prediction model.
One direction of work aims to compute attribute importance by querying the prediction model with randomly generated samples. 
Many of the techniques leverage the popular technique \lime\ introduced by Ribeiro et al.~\cite{ribeiro2016should}. 
\lime\ generates a perturbation of samples around the instance to be explained.
Then, using the generated samples and their corresponding black-box predictions, \lime\ trains a locally faithful model that can approximate the local decision boundary of a prediction model.
Rehse et al. apply \lime\ to a smart production process, where an LSTM is trained to solve a binary classification problem~\cite{DBLP:journals/ki/RehseMF19}. 
Sindhgatta et al. apply \lime\ to interpret the results of eXtreme Gradient Boosting trees applied for outcome-oriented and time-oriented prediction problems~\cite{SindhgattaOM20}. 
Rizzi et al. use \lime\ to understand the weakness of random forest classifiers for outcome-oriented predictions~\cite{DBLP:conf/bpm/RizziFM20}. 
They find the most frequent attributes that are identified as important by \lime\ when the classifier makes wrong predictions (false positives and false negatives). Then, they randomize these attributes to train the model again to neutralize the effects. 
Despite its popularity, one concern is that \lime{}'s sampling process ignores possible correlations among the input attributes \cite{molnar2019} e.g., process constraints in the case of predictive process monitoring \cite{jan2020ai}.
Thus, unrealistic synthetic instances might be used to train a local interpretable model, thereby leading to unrealistic explanations.
Another sampling-based approach is followed by Galanti et al., who use Shapley Values (from game theory) to provide explanations to the users~\cite{GalantiCLCN20}.
They visualize explanations as an aggregated heatmap of attribute importance summarizing all the cases in the test set. Another direction of work involves Partial Dependence Plots, which are used by Mehdiyev and Fettke to derive the marginal effects between attributes and predictions from a neural network~\cite{MehdiyevF20}. 

Again, the above work focuses on factual explanations only.
By generating counterfactual explanations, our technique complements this work.

%% file: prob.tex
\section{Motivation and Problem Statement}
\label{sec:rationale}

Our technique is based on the \lore\ technique introduced by Guidotti et al.~\cite{DBLP:journals/expert/GuidottiMGPRT19}. 
\lore\ is a local model-agnostic technique that uses genetic algorithms to generate synthetic data for training decision trees that locally approximate the behavior of a black-box prediction model. 
\lore\ uses genetic algorithms to iteratively generate a set of neighboring input data instances, which are in close proximity to the instance to be explained. 

Using decision trees as the interpretable models offer two main advantages:
First, decision trees are intrinsically interpretable models.
Second, explanations in the form of rules (factual and counterfactual) can be generated from a decision tree by following the split conditions from the root to the leaf node \cite{DBLP:journals/expert/GuidottiMGPRT19}. 

These rule-based explanations provide a step forward from simply generating a single counterfactual data instance as an explanation~\cite{wachter2017counterfactual}, as rules provide a higher level of abstraction of the explanation\footnote{If concrete counterfactual data instances are needed, these can be generated, e.g., by randomly sampling concrete values for the attributes that comply with the rules.}. 

However, directly applying \lore\ to prediction models for predictive process monitoring faces two problems:

\runin{Problem 1.} Perturbation-based techniques, like \lime\ and \lore{}, generate random samples without considering existing process constraints \cite{jan2020ai}.
Typically, there are constraints in which order the activities in a business process may be executed imposed by the control flow.
As an example, activity B may only follow after activity A was executed. 
As a result, such process-constraint-agnostic perturbation-based techniques face the risk of generating unrealistic instances that represent process traces that could never happen. 
In turn, the generated interpretable models deliver misleading or useless explanations; e.g., the counterfactual explanation may indicate that executing a specific activity B before A may lead to the desired outcome, while such execution order is not possible.

\runin{Problem 2.} \lore\ is limited to binary classification. 
In BPM applications, however, one very often faces multi-class prediction problems.
Examples include next activity predictions~\cite{TaxVRD17}, where the classes are the number of activity types that may occur, and outcome-oriented predictions, where more than two process outcomes may be predicted~\cite{TeinemaaDRM19}.

%% file: approach.tex
\section{Approach}
\label{sec:approach}

To address the above problems, we introduce the \loreley\ technique, which offers several enhancements from \lore{}, thereby delivering a specific technique for generating counterfactual explanations for predictive process monitoring.
In the following, we first briefly explain \lore\ and then introduce \loreley{}.

\subsection{The LORE Technique as Starting Point}
\label{sec:details}

Given a black-box prediction model $b$, and an input instance to be explained $x$, \lore\ provides an explanation for the black-box prediction $b(x) = \hat{y}$. 
\lore\ consists of two main stages: (1) neighborhood generation, and (2) interpretable model induction and explanation generation~\cite{DBLP:journals/expert/GuidottiMGPRT19}. 

\runin{Stage 1.} In this stage, a group of data instances $Z$ is generated using a genetic algorithm.
The goal of the genetic algorithm is to a generate a balanced neighborhood $Z = Z_{=} \cup Z_{\neq}$, where instances $z \in Z_{=}$ satisfy $b(x) = b(z)$ and instances $z \in Z_{\neq}$ satisfy $b(x) \neq b(z)$.
To initiate the first generation $P_0$ for the genetic algorithm, Guidotti et al. propose making $N$ copies of $x$ as the initial generation. 
Following the initialization step, generation $P_0$ goes through the evolutionary loop.
The loop starts with selecting a group $A$ from $P_i$ with the highest fitness score according to the fitness function. 
Then, the crossover operation is applied to $A$ with a probability of $p_c$. 
For the crossover operation, two-point crossovers are used to select two attributes from the parents randomly and swap the selected attributes to generate the children.
The resulting instances and the ones without crossover are placed in a set $B$. 
Then, a proportion of $B$ is mutated with probability $p_m$.
The attributes are randomly replaced by the values generated by the empirical distribution derived from the data \cite{DBLP:journals/expert/GuidottiMGPRT19}. 
Finally, The unmutated and mutated instances become the next generation $P_{i+1}$ and go through the next evolutionary loop.
These evolutionary loops are repeated until the user-defined generation $G$ is reached.
The genetic generation process is run for both $Z_{=}$ and $Z_{\neq}$ to form the neighborhood $Z$ of $x$. 

\runin{Stage 2.} After the neighborhood generation stage, an interpretable model $c$ in the form of a decision tree classifier is induced to derive explanations for $x$, the instance to be explained. 
Once the decision tree classifier is trained, the factual rule for the instance to be explained $x$ can be derived by following the split conditions of the tree from root to the leaf where $x$ belongs.
As for the counterfactual rule, it is defined as the rule derived from the path leading to a different outcome with the minimum number of violated split conditions (not satisfied by $x$) on the path. 
One advantage of using a decision tree as the interpretable model $c$ is that both factual and counterfactual explanations in the form of decision rules are consistent with $c$ by construct \cite{DBLP:journals/expert/GuidottiMGPRT19}.
Also, one can tune how detailed and precise the explanation should be by setting the complexity of the tree (e.g. max depth, minimum impurity decrease, etc.).

\subsection{Preliminaries} \label{sec:prelim}
Based on \cite{TeinemaaDRM19,DBLP:journals/tist/VerenichDRMT19}, we define key concepts related to predictive process monitoring:

\begin{itemize}
    \item \textbf{Event:} An event $e$ is a tuple $(a,c,t,(d_1, v_1), ...,(d_n, v_n))$, where $a$ is the event class, $c$ is the case id, $t$ is the timestamp, $(d_1, v_1)....(d_n, v_n), n \geq 0$ are the name and value of the event attributes.
    \item \textbf{Trace}: A trace is a sequence $ \sigma = \langle e_1, e_2, ..., e_m, \{(b_1, w_1), ...,(b_s, w_s)\}\rangle $, $\forall i, j \in \{1,\ldots,m\}, e_i.c = e_j.c$, which means all events in a trace carry the same case id. $\{(b_1, w_1), ..., (b_s, w_s)\}, s \geq 0$ are the name and value of the case attributes.
    \item \textbf{Event log}: An event log is a set of completed traces. $L = \{\sigma_1, \sigma_2, ..., \sigma_g\}$, where $g$ is the number of traces in the event log.
    \item \textbf{Trace prefix}: $\sigma|_k = \langle e_1, e_2, ..., e_k\rangle$ is the prefix of length $k$ (with $0 < k < m$) of trace $\sigma$; e.g., $\langle e_1, e_2, e_3, e_4\rangle|_2 = \langle e_1, e_2\rangle$.
\end{itemize}

\subsection{The LORELEY Technique}
\label{sec:loreley}

\begin{figure}[h]
	\includegraphics[width=\textwidth]{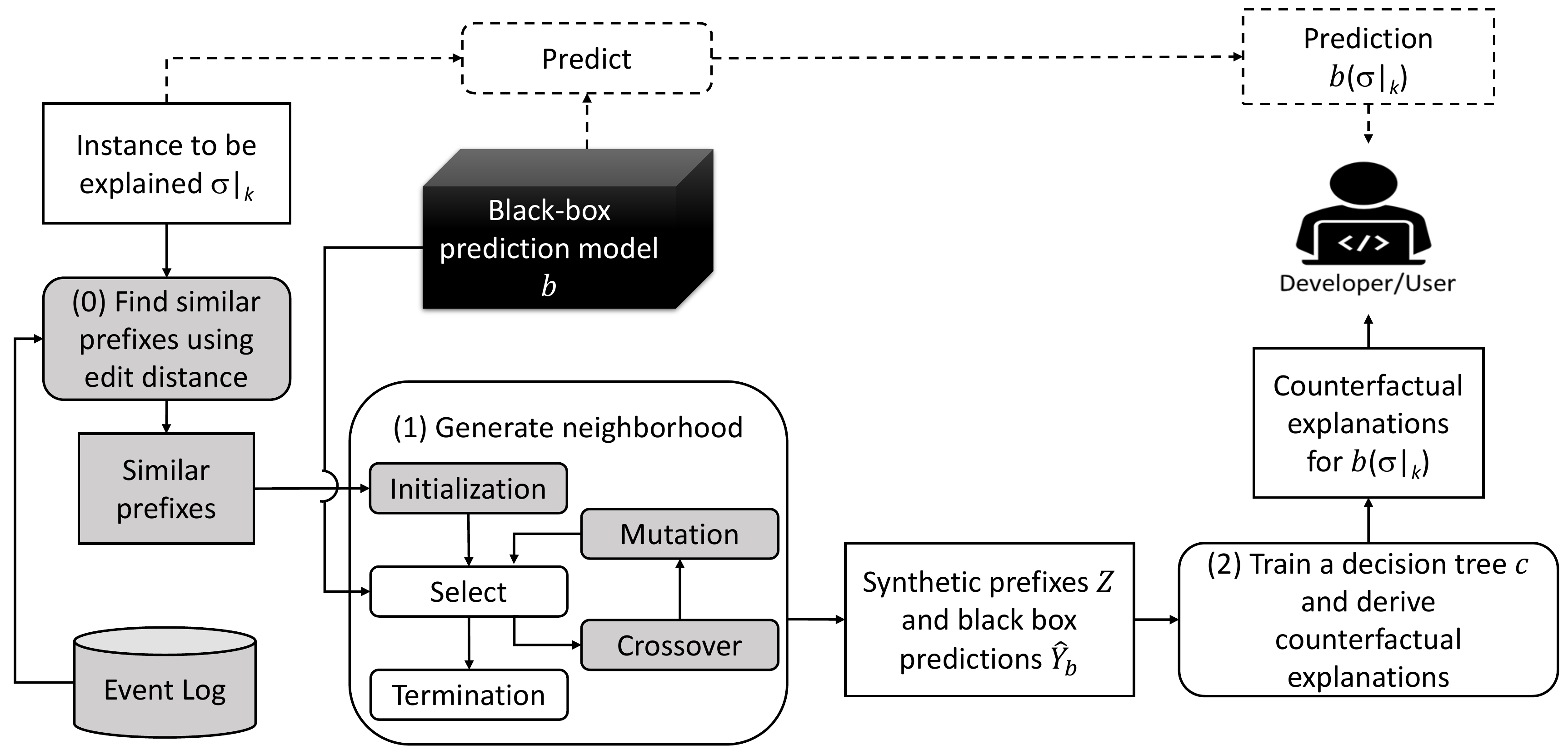}
	\caption{Overview of \loreley\ (extensions from \lore\ highlighted in grey)}
	\label{fig:approach}
\end{figure}

\loreley\ includes three extensions of \lore\ that jointly address the two problems as motivated in Sec.~\ref{sec:rationale}. 
A graphical overview of the artifacts and activities of \loreley\ is shown in Fig. \ref{fig:approach}.
The upper part of the figure depicts how individual predictions are generated. 
As this is not the focus of \loreley{}, it is depicted by dashed lines.
The lower part of the figure depicts how explanations are generated and highlights the extensions from \lore\ in gray.
This consists of three main stages: Stage 0 is an additional stage compared with \lore\ and is responsible for finding similar prefixes for initialization.
Stage 1 generates neighborhood instances using a modified form of the genetic algorithm from \lore{}.
Stage 2 follows \lore{} to train a decision tree for deriving counterfactual explanations.

\runin{Initialization with Similar Prefixes.}
To ensure that mutation possibilities for the control flow come from real traces, we modify the initialization step of \lore\ as follows.
While \lore{} duplicates the instance to be explained to reach the target population size for the first generation, \loreley{} selects similar prefixes in the event log by comparing the edit distance (see below) on the control flow sequences, thereby incorporating more variation in the first generation. 
This is key because the control flow variations in the first generation are used during the mutation step of the genetic algorithm.

The concept of edit distance was used by \cite{DBLP:journals/tsc/Francescomarino19,DBLP:conf/caise/MaggiFDG14} to group similar traces during trace bucketing as part of encoding the input data for the prediction model. 
The assumption behind such bucketing is that prefixes with similar control flow are more likely to have similar behaviors in the future. 
We follow this reasoning and adopt the Levenshtein edit distance to find prefixes that are similar to the instance to be explained (in terms of control flow).

\runin{Considering Control Flow Constraints during Crossover \& Mutation.}
During the crossover and mutation steps, \loreley\ considers additional constraints related to attributes that represent control flow. 
In many predictive process monitoring models, the control flow is represented as a combination of several one-hot encoded categorical attributes or as aggregations (e.g., frequency) of the control flow attributes \cite{TaxVRD17,TeinemaaDRM19}. 
Without further constraints, crossover and mutation would treat each of these encoded attributes independently without considering the potential constraints implied by the underlying process.
As a result, crossover and mutation would lead to instances that represent unrealistic control flows.

\begin{figure}[h]
    \centering
	\includegraphics[width=.9\textwidth]{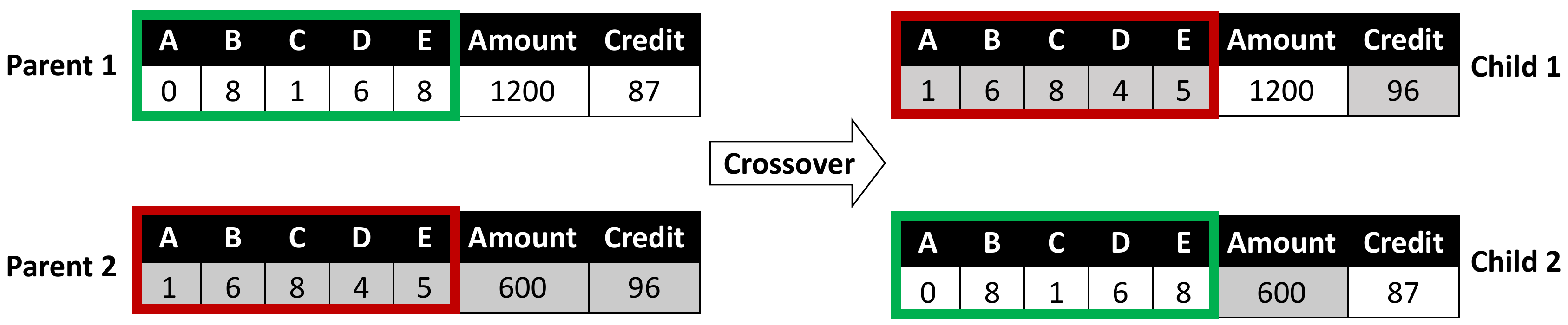}
	\caption{Example for control-flow-aware crossover in \loreley{}}
	\label{fig:extension_example}
\end{figure}

To address this problem, we propose to treat the attributes representing the control flow as a single attribute in the crossover and mutation steps.
Fig. \ref{fig:extension_example} shows an example.
In the crossover step, the attributes representing the control flow (encoded as frequency vectors in the example) are combined and swapped with other bundles of control flow attributes from the other parent to produce children.
In the mutation step, the possible control flow attributes are replaced by the values randomly drawn from the control flow attributes of the first generation.
By doing so, we ensure that the control flows of the synthetic prefixes are similar to the instance to be explained $x = \sigma|_k$ (within the user-defined similarity threshold) and are realistic by construction.

\runin{Explaining Multi-Class Predictions.}
We extend \lore\ from binary prediction to multi-class prediction: The neighborhood becomes 
$\label{new_neigh}
 Z = \bigcup_{i \in I} Z_{b(z) = i}
$,
where $I$ is the set of possible predicted outcomes and $z$ are the instances with black-box prediction $b(z) = i, i \in I$. 
To generate each $Z_{b(z) = i}$, the fitness function of \lore{} is extended to
\begin{equation}
    \label{eq:fitness_function}
    fitness_{i}^{x}(z) = \mathbbm{1}_{b(z) = i} + (1 - d(x, z)) - \mathbbm{1}_{x = z}
\end{equation}
where $d$ is a distance function, $\mathbbm{1}$ is an indicator function $\mathbbm{1}_{true} = 1, \mathbbm{1}_{false} = 0$.
The intuition behind is to find the instance $z$ that is as close as possible to the instance to be explained $x$, yet not equivalent, while the corresponding black-box prediction $b(z)$ should be equal to a predefined outcome, i.e. $b(z) = i$ \cite{DBLP:journals/expert/GuidottiMGPRT19}.

%% file: evaluation.tex
\section{Experiment Setup}
\label{sec:Experiments}

To facilitate transparency, reproducibility, and replicability, the code of our prototypical implementation, the used dataset as well as our experimental results are publicly available\footnote{\scriptsize \url{https://git.uni-due.de/adi645f/cf4bpm-artifacts}}. 


\runin{Research Questions.} We aim to answer the following  research questions:

\emph{RQ1.} \emph{How faithful are the explanations generated by \loreley\ to the underlying black-box prediction model?} We analyze how accurately the interpretable model $c$ can approximate the local decision boundary of the black-box model $b$.

\emph{RQ2.} \emph{How is the quality of \loreley{}'s explanations?} We analyze how far the generated explanations match what would be expected considering domain knowledge and insights about the process.

\runin{Data Set.}
We use the BPIC2017 event log\footnote{\scriptsize \url{https://doi.org/10.4121/uuid:5f3067df-f10b-45da-b98b-86ae4c7a310b}}, which is widely used in  predictive process monitoring research~\cite{TeinemaaDRM19,DBLP:journals/tist/VerenichDRMT19}.
The BPIC2017 event log concerns the loan application process of a Dutch financial institute and covers 31,509 cases. 
The events are structured into three types: events that record \textbf{A}pplication state changes, events that record \textbf{O}ffer state changes, and \textbf{W}orkflow events. 
The overall application process can be decomposed into three stages, receiving applications, negotiating offers, and validating documents respectively \cite{rodrigues2017stairway}.
Each stage has its corresponding working items (workflow events) and the application state is updated according to the stages that the application has gone through.
At the end of the process, an application could be successful ($A\_Pending$)\footnote{\scriptsize The assessment is positive, the loan is final and the customer is paid.} or not ($A\_Denied$).
Additionally, an application state $A\_Canceled$ is set if the customer neither replied to the call nor sent the missing documents as requested.

\runin{Training the Black-Box Prediction Model.}
We train a black-box prediction model $b$, for which we generate explanations.
The model $b$ predicts the process outcome at different prediction points (i.e., prefix lengths).
Following~\cite{rodrigues2017stairway,TeinemaaDRM19}, we consider the occurrence of the three events, A\_Pending, A\_Canceled, and A\_Denied as the respective process outcomes.

\begin{wrapfigure}[11]{r}{.5\textwidth}
		\centering
		\vspace{-2em}
			\includegraphics[width=.43\textwidth]{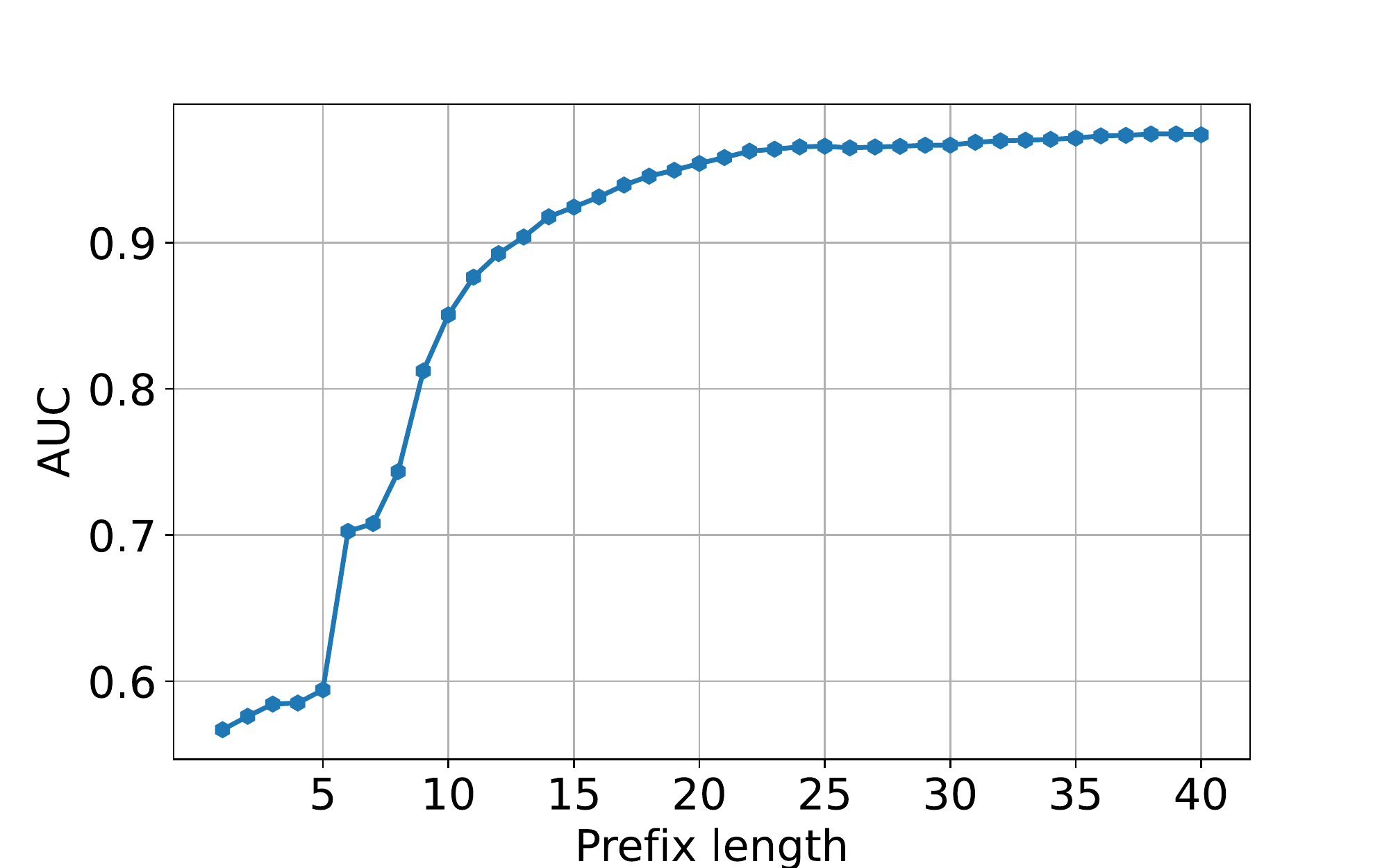}
		\caption{AUC by different prefix lengths}
		\label{fig:auc}
\end{wrapfigure}

We train a two-layer LSTM model, a widely used black-box prediction model for predictive process monitoring (e.g., see~\cite{CamargoDG19,EvermannRF17,TaxVRD17}).
We split the event log into a training set (80\% of the traces) and a testing set (20\% of the traces).
As we are not aiming at improving the prediction accuracy of the black box with \loreley{}, we did not perform exhaustive hyper-parameters/structures tuning.

Following Teinemaa et al.~\cite{TeinemaaDRM19},  we measure the prediction accuracy of the LSTM using the area under the ROC curve (AUC) metric.
Fig.~\ref{fig:auc} shows  AUC across different prefix lengths.
The overall AUC is 0.857, which is calculated as the weighted average of the obtained AUC for each prefix length up to 40.



\runin{Generating Explanations.}
As the prediction accuracy of the trained black-box model from above differs for the different prefix lengths, this allows us to evaluate \loreley\ in different scenarios.
In particular, we can evaluate our technique when the black-box model delivers low accurate predictions and also when predictions are getting more and more accurate.
As such, we evaluate \loreley{} for prefix lengths 5, 10, 20, 30, and 40. 
As shown in Fig.~\ref{fig:auc}, there is a jump in AUC between prefix lengths 5 and 6. 
After that, accuracy gets constantly higher.
For each prefix (in the test set) belonging to the aforementioned prefix lengths, we generate an interpretable model $c$ using \loreley{}. 
This means we generate a total of 23,426 interpretable models (see Table~\ref{tab:results_metrics} below).

As parameters for the genetic algorithm, we use the suggested default values from~\cite{DBLP:journals/expert/GuidottiMGPRT19} with some small adjustments. 
We use default values for crossover and mutation probabilities, $p_m=0.7$ and $p_c=0.2$, but increase the number of generations $G$ from 10 to 15 due to the highly imbalanced dataset (Pending: 55\%, Canceled: 33\%, Denied: 12\%).
The fitness calculation (see Eq.~\ref{eq:fitness_function}) has to query the black-box model to retrieve the corresponding prediction for every generated prefix.
To keep the experiment computationally tractable, we set a target population size of 600 for each of the three possible outcomes, resulting in 1800 prefixes $Z$ to train and evaluate the explanation. 
The generated prefixes are split into 80\% training set and 20\% test set. 
The interpretable model $c$ (i.e., a decision tree) is trained using the training set and evaluated on the test set. 

\section{Experiment Results}
\label{sec:results}

\runin{Results for RQ1 (Faithfulness of Explanations).}
We use the widely used fidelity metric from explainable AI \cite{DBLP:journals/expert/GuidottiMGPRT19,guidotti2018survey,ribeiro2016should} to evaluate the faithfulness of \loreley{}.
We evaluate how faithful the interpretable model $c$ and thus the explanations generated from it are to the black-box model $b$.



Table \ref{tab:results_metrics} reports the evaluation results.
The average results reported are calculated as weighted average considering the number of prefixes per prefix length in the test data. 
Overall, the interpretable model generated by \loreley{} reaches 97.69\% $Fidelity$ for all the evaluated prefix lengths on average. 
$Fidelity$ remains stable across all prefix lengths with a variance of less than 0.5\%, suggesting that \loreley\ can generate good explanations independent of the accuracy of the black-box model.

\begin{table}[h]
	\centering
	\resizebox{.65\textwidth}{!}{%
	\arrayrulecolor{black}
	\begin{tabular}{c|c|c} 
  \emph{Prefix Length }& \emph{Nbr. of Prefixes }    & $Fidelity$   \\ 
	\hline
	5                                                        & 6,283       & 0.9754  \\
	10                                                       & 6,277       & 0.9735  \\
	20                                                       & 5,338       & 0.9849  \\
	30                                                       & 3,454       & 0.9755  \\
	40                                                       & 2,074       & 0.9730  \\ 
	\hline
	& \textbf{Total: 23,426} & \textbf{Avg.: 0.9769}  \\
	\end{tabular}}
	\vspace{.5em}
	\caption{Metrics for different prefix lengths}
	\label{tab:results_metrics}
	\arrayrulecolor{black}
\end{table}
The results show that \loreley{} can approximate the black-box model's local decision boundary accurately.
This implies that the explanations generated by \loreley{} can faithfully represent the knowledge learned by the black-box model within the local decision boundary for the instance to be explained.

\runin{Results for RQ2 (Quality of Explanations).}
To assess the quality of explanations generated by \loreley{}, we look deeper into its explanations. 
First, we examine whether the most important attributes indicated by the explanations correspond to the domain knowledge. 
It is expected that the logic learned by the black box $b$ would reflect a certain degree of domain knowledge as accuracy increases along the prefix length.
Since involving the process owner of the BPIC2017 event log was not possible, we compare our results with the findings of the winners of the BPIC2017 challenge, for which this dataset was published.

\begin{figure}[t!]
	\centering
	\includegraphics[width=\textwidth]{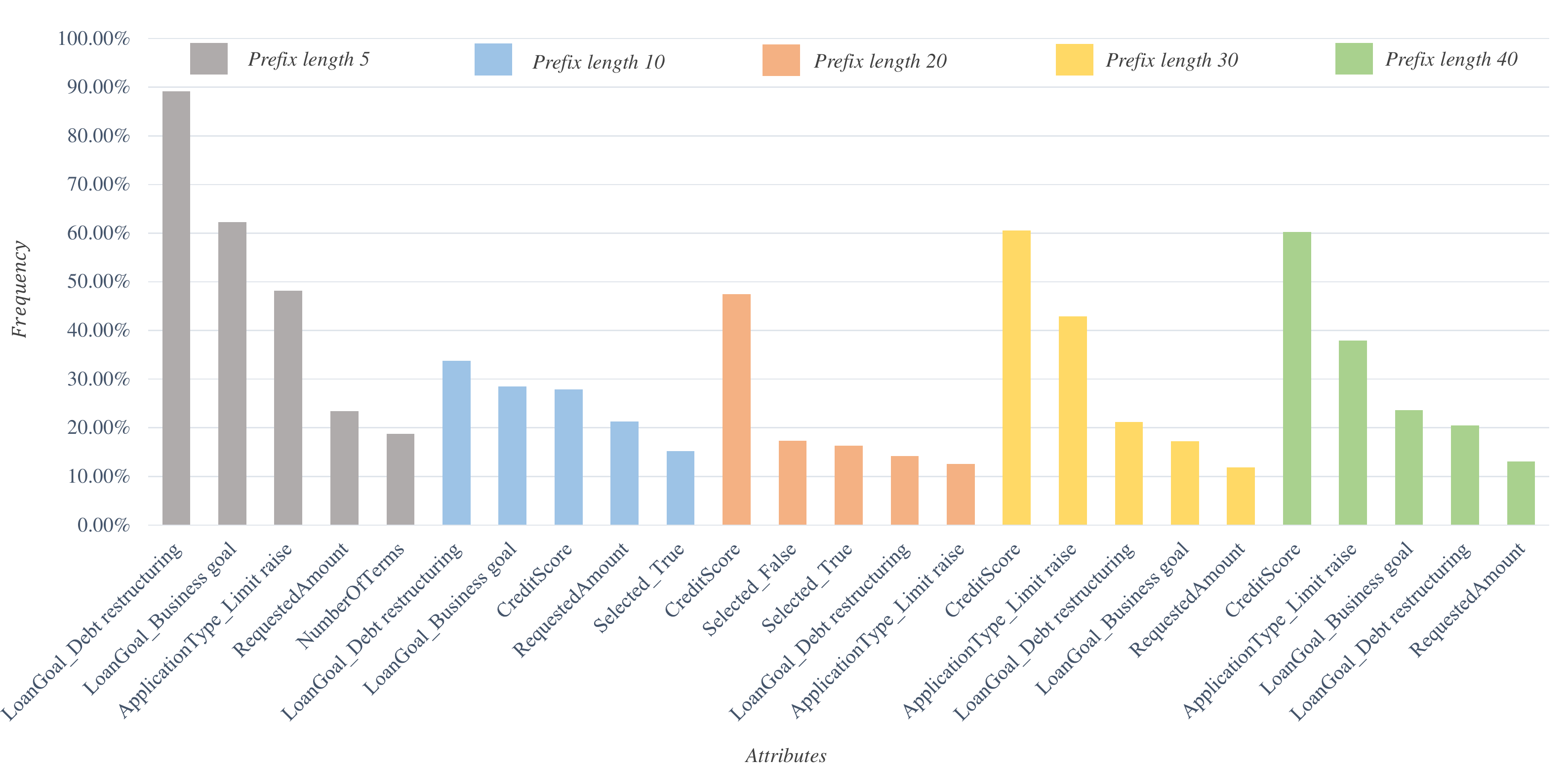}
	\caption{Top five frequent important attribute for different input length} 
	\label{fig:most_frequent_attributes}
\end{figure}

Fig. \ref{fig:most_frequent_attributes} shows the top five most frequent attributes for each input length examined.
The most important attributes can be computed from the decision tree according to the decrease of impurity used to select split points.
For every interpretable model $c$, the most important attributes are tracked. 
Then, the attributes are counted, grouped by the prefix length, and divided by the number of instances of each prefix length.
The result shown in Fig.~\ref{fig:most_frequent_attributes} gives a broader view regarding what the black box $b$ has learned for different input lengths.
As the figure shows, the most important attribute for prefix lengths 20, 30, and 40 is $CreditScore$. 
This corresponds to the findings of the professional group winners~\cite{blevi2017process}, where they found $CreditScore$ is among the most important attributes as a result of the predictive analysis.
The other frequent attribute is \emph{ApplicationType\_Limit raise} and it is the second most frequent important attribute for prefix length 30 and 40.
This corresponds to the findings of the academic group winners~\cite{rodrigues2017stairway}, who found that the institute approved a significantly higher fraction of applications of the type ``limit raise'' (73.37\%) than ``new credit'' (52.61\%).
It is therefore another reasonable predictor used by the black box $b$.
Fig. \ref{fig:most_frequent_attributes} also shows that the \emph{LoanGoal} attributes are used by the black-box model quite often to predict the outcome (especially for prefix lengths 5 and 10) although \emph{LoanGoal} has little influence on the outcome of the case according to the analysis \cite{rodrigues2017stairway}.
However, they could be the most frequent decisive attributes available for the black box considering prefixes 5 and 10 are still quite early in the process.

Complementing this high-level assessment of explanation quality, we analyze the explanations generated by \loreley{} for a particular input data of prefix length 30. 
The instance to be explained has the following event attributes for \emph{O\_Create Offer}  
\{\emph{FirstWithdrawalAmount }= 0,
	\emph{NumberOfTerms }= 126,
	\emph{MonthlyCost }= 250,
	\emph{CreditScore }= 0,
	\emph{OfferedAmount }= 25000,
	\emph{Selected }= true,
	\emph{accepted }= false\}, and
	it has the following case attributes 
	\{\emph{case\_RequestedAmount }= 25000, 
	\emph{case\_LoanGoal }= \emph{Existing loan take over}, 
	\emph{case\_ApplicationType }= \emph{new credit}\}.
The corresponding black-box prediction is $A\_Denied$.
By observing the control flow of the prefix, one can find that the prefix is at the later stage of the application process. 
The application process has gone through application creation and assessment. 
An offer is created and the customer has been contacted a few times to send incomplete files.

\begin{figure}
	\centering
	\includegraphics[width=0.8\textwidth]{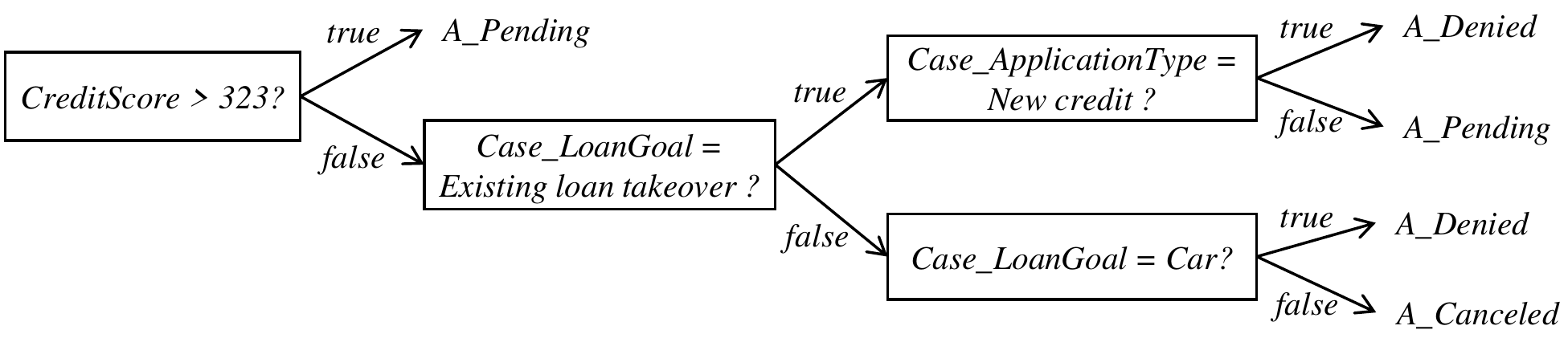}
	\caption{An example of a generated interpretable model $c$} 
	\label{fig:dt_example}
\end{figure}

For this instance to be explained, \loreley{} generates the interpretable model $c$ shown in Fig.~\ref{fig:dt_example}.
The $Fidelity$ of $c$ is 0.9343.

The factual rule for the instance to be explained is: 

\vspace{-.5em}
\begin{center}
	\{\emph{CreditScore } $\leq$ 323, \emph{Case\_LoanGoal }= \emph{Existing loan takeover}, \\
	\emph{Case\_ApplicationType }= \emph{New credit}\} $\rightarrow$ \emph{A\_Denied }
\end{center}
\vspace{-.5em}

The following are examples of two counterfactual rules (with changes from the factual rule highlighted in \hl{grey}):

\vspace{-.5em}
\begin{center}
	\{\emph{CreditScore} \hl{$>$} 323\} $\rightarrow$ \hl{\emph{A\_Pending}}
\end{center}

\vspace{-1.5em}
\begin{center}
	\{\emph{CreditScore} $\leq$ 323, \emph{Case\_LoanGoal }= \emph{Existing loan takeover}, \\
	\emph{Case\_ApplicationType} \hl{!=} \emph{New credit}\} $\rightarrow$ \hl{\emph{A\_Pending}}
\end{center}
\vspace{-.5em}

These two counterfactual rules provide the least changes (split conditions not fulfilled by the data instance to be explained) one has to make to obtain the desired outcome.
Again, the derived explanations match the findings of the BPIC2017 academic winner \cite{rodrigues2017stairway}: Applications of type ``limit raise'' or applicants with higher credit scores indeed have higher acceptance rates (recall that state $A\_Pending$ means accepted; see above).

\runin{Threats to Validity.}
Concerning external validity, we used a large, real event log (BPIC2017) widely used in related work.
Still, we only performed our experiments for a single event log and used only one type of black-box prediction model, which thus limits the generalizability of our findings.

With respect to internal validity, sample-based explanation techniques (including \lime{}, \lore{}, and \loreley{}) face potential instability due to their stochastic nature.
That is, the generated explanations can be expected not to be the same when applying the techniques to the same input. 
As a result, the metrics reported above might be different.
Examining the stability of the technique requires significant additional compute resources and time.

%% file: limitations.tex
\section{Limitations}
\label{sec:limit}

\runin{Similarity threshold depends on the encoding of prefixes.}
Since the fitness calculation in Eq. \ref{eq:fitness_function}, specifically the distance calculation $d(x, z)$, needs the attribute vectors to be the same length, the similarity threshold depends on how the prefix is encoded.
For example, when using index-based encoding~\cite{TeinemaaDRM19}, setting the similarity threshold to $0$ may be necessary because the fitness function cannot compute the distance between two prefixes with different lengths.
This dependency is due to the distance calculation in the fitness function.
A way to avoid this is to map the prefix vectors of different lengths to the same dimensions for fitness calculation by using dimensionality reduction techniques.

\runin{Finding influential predictors based on control flow.} Another limitation of \loreley{} is that it may not be able to find influential predictors that are based on control flow. 
We use edit distance to identify prefixes that have similar control flow. 
However, two prefixes with very large edit distance could be very ``similar'' based on a process model. 
Take BPIC2017 as an example, we might want to explain an instance with only one offer event in the prefix. 
While \cite{rodrigues2017stairway} shows that the number of offers is one of the decisive factors to the outcome of the process, \loreley{} is unlikely to include prefixes with more than one offer due to the way it calculates similar prefixes.
A possible extension of \loreley{} could be to use the process model to identify similar prefixes instead of using the edit distance.